%% file: main.tex
\documentclass{article}

\usepackage{fullpage}

\usepackage[utf8]{inputenc} 
\usepackage[T1]{fontenc}    
\usepackage{lmodern}
\usepackage{booktabs}       
\usepackage{amsfonts}       
\usepackage{nicefrac}       
\usepackage{microtype}      
\usepackage{color}

\usepackage{amsmath}
\usepackage{amssymb}
\usepackage{algorithm2e}

\usepackage{tikz}
\usetikzlibrary{positioning,decorations.pathreplacing}

\usepackage{wrapfig}
\usepackage{floatflt}

\newcommand{\sign}{\mathrm{sign}}

\newcommand{\bb}{{\bf b}}

\newcommand{\be}{{\bf e}}

\newcommand{\bh}{{\bf h}}

\newcommand{\br}{{\bf r}}
\newcommand{\bs}{{\bf s}}

\newcommand{\bu}{{\bf u}}
\newcommand{\bv}{{\bf v}}
\newcommand{\bw}{{\bf w}}
\newcommand{\bx}{{\bf x}}

\newcommand{\R}{{\bf R}}
\newcommand{\hy}{\hat{y}}

\renewcommand{\Pr}{{\bf Pr}}

\newcommand{\bdelta}{{\boldsymbol{\delta}}}
\newcommand{\cW}{ {\cal W} }
\newcommand{\cD}{ {\cal D} }
\newcommand{\cR}{ {\cal R} }

\newcommand{\Wsquare}{ || { \cW } ||_2^2} 
 
\newcommand{\eps}{\epsilon}

\newcommand{\qed}{\hfill\rule{7pt}{7pt} \medskip}

\DeclareSymbolFont{bbold}{U}{bbold}{m}{n}
\DeclareSymbolFontAlphabet{\mathbbold}{bbold}
\newcommand{\ind}{ { \mathbbold{1} } }

\newcommand{\Bin}{ {\mathrm{Bin}} }

\newcommand{\E}{{\bf E}}
\newcommand{\Var}{{\bf Var}}

\newcommand*{\defeq}{\stackrel{\textup{def}}{=}}

\newcommand{\relunode}{  
    \begin{tikzpicture}
      \draw[thick, red] (-0.2,0) -- (0,0) -- (0.15, 0.15) ;
    \end{tikzpicture}
}
\newcommand{\linnode}{  
    \begin{tikzpicture}
      \draw[thick, red] (0,0) -- (0.28, 0.28) ;
    \end{tikzpicture}
}
\tikzstyle{neuron}=[draw, circle, thick, black, text=red]  

\newtheorem{theorem}{Theorem}

\newtheorem{lemma}[theorem]{Lemma}

\newtheorem{definition}[theorem]{Definition}

\title{\bf Surprising properties of dropout \\
           in deep networks \\}

 \author{David P. Helmbold\\
       Department of Computer Science\\
       University of California, Santa Cruz \\
       Santa Cruz, CA 95064, USA \\
       \texttt{dph@soe.ucsc.edu} \\
 \and
  Philip M. Long \\
  Google \\
  \texttt{plong@google.com}\\
}

\begin{document}

\maketitle

\begin{abstract}

We analyze dropout in deep networks with rectified linear units and the quadratic loss.
Our results expose surprising differences between 
the behavior of dropout and more traditional regularizers like weight decay.
For example, on some simple data sets dropout training produces negative weights even though the output is the sum of the inputs.
This provides a counterpoint to the suggestion that dropout discourages co-adaptation of weights.
We also  show that the dropout penalty can grow exponentially in the depth of the network
while the weight-decay penalty remains essentially linear, and that dropout is insensitive to 
various re-scalings of the input features, outputs, and network weights.
This last insensitivity implies that there are no isolated local minima of the dropout training criterion.
Our work uncovers new properties of dropout, extends our understanding of why dropout succeeds,
and lays the foundation for further progress.
%
%
%

\end{abstract}

\input{intro}

\input{prelim}

\input{scale}

\input{negative_weights}

\section{Properties of the dropout penalty}
\label{s:dropout.penalty}

\input{growth}

\input{negative_penalty}

\input{depends}

\input{experiments}

 \input{conclusions}

\section*{Acknowledgments}

We are very grateful to Peter Bartlett, Seshadhri Comandur, 
and anonymous reviewers for valuable communications.

\small
\bibliographystyle{plain}
\bibliography{general}

\normalsize

\appendix
\input{notation_table}

\input{elltwo}

\end{document}

%% file: intro.tex
\section{Introduction}
\label{s:intro}

The 2012 ImageNet Large Scale Visual Recognition challenge was 
won by 
the University of Toronto team by a surprisingly large margin.
In an invited talk at NIPS, Hinton \cite{Hin12} credited the dropout training technique for 
much of their success.
Dropout training is a variant of stochastic gradient descent (SGD) where, as each example is processed, 
the network is temporarily perturbed by randomly ``dropping out'' nodes of the network.
The gradient calculation and weight updates are performed on the reduced network, 
and the dropped out nodes are then restored before the next SGD iteration.
Since the ImageNet competition, dropout has been successfully applied to a variety of 
domains  
\cite{Dah12,DenEtAl13,DSH13,Kal14,chen2014fast},
and is widely used \cite{schmidhuber2015deep,he2015delving,Szegedy_2015_CVPR}; 
for example, it is incorporated into popular
packages such at Torch \cite{Tor16}, Caffe \cite{Caf16} and TensorFlow
\cite{Ten16}.  
It is intriguing that crippling the network during training often leads to such dramatically improved results,
and dropout has also sparked substantial research on
related methods (e.g. \cite{GooEtal13,WanEtAl13}).

In this work, we examine the effect of dropout on the inductive bias
of the learning algorithm.
A match between dropout's inductive bias and
some important applications could explain
the success of dropout,
and its popularity also motivates the study of its unusual properties.

Weight decay training optimizes the empirical error plus an $L_2$ regularization term, 
$\frac{\lambda}{2} ||\bw||_2^2$, so
we call $\frac{\lambda}{2} ||\bw||_2^2$ the 
\underline{\emph{$L_2$ penalty}} of $\bw$ since it is the difference between 
training criterion evaluated at $\bw$ and the empirical loss of $\bw$.
By analogy, we define the \underline{\emph{dropout penalty}} of $\bw$ to be the difference between the dropout training criterion  
and the empirical loss of $\bw$  (see Section~\ref{s:prelim}).
Dropout penalties measure how much dropout discriminates against weight vectors, so they are key
to understanding dropout's inductive bias.
%

%
Even in one-layer networks, conclusions drawn from (typically quadratic) approximations of the
dropout penalty can be misleading \cite{HL15}.
Therefore we focus on exact formal analysis of dropout in multi-layer networks.
%
%
%
Theoretical analysis of deep networks is notoriously difficult, so we might expect that
a thorough understanding of dropout in deep networks must be achieved in stages.  
In this paper we further the process by exposing some of the surprising ways that the
inductive bias of dropout differs from $L_2$ and other standard regularizers.    
These include the following:

\begin{itemize}
\item We show that dropout training can lead to negative weights \emph{even when the output is a positive multiple of the the inputs.}  Arguably, such use of
  negative weights constitutes co-adaptation -- 
  this adds a counterpoint
  to previous analyses showing
  that dropout discourages co-adaptation \cite{SriEtAl14,HL15}.
\item Unlike weight decay and other $p$-norm regularizers, dropout training is insensitive to the rescaling of input features, and largely insensitive to rescaling of the outputs;  this may play a role in dropout's practical success.
Dropout is also unaffected if the weights in one layer are scaled up by
a constant $c$, and the weights of another layer are scaled down by $c$;
this implies that dropout training does not have isolated local minima.
\item The dropout penalty grows exponentially in the depth of the network in cases where the $L_2$ regularizer grows linearly.
  This may enable dropout to
  penalize the complexity of the network in a way that more
  meaningfully   reflects the richness of the network's  behaviors.
(The exponential growth with $d$ of the dropout
penalty is reminiscent of some regularizers for deep networks studied
by Neyshabur, et al \cite{NTS15}.)
\item Dropout in deep networks has a variety of other behaviors different from standard regularizers. 
In particular:  the dropout penalty for a set of weights can be negative;
 the dropout penalty of a set of weights depends on both the training instances and the labels;
and although the dropout probability intuitively measures the strength of dropout regularization, 
the dropout penalties are often non-monotonic in the dropout probability.
In contrast, Wager, et al \cite{WWL13} show that when dropout is applied to 
generalized linear models,
the dropout penalty is always non-negative and does not depend on the labels.
\end{itemize}




Our analysis is for multilayer neural networks with the square loss at the output node.  
The hidden layers use 
the popular rectified linear units 
\cite{nair2010rectified}
outputting $\sigma(a) = \max(0, a)$ where $a$ is the node's activation (the weighted sum of its inputs).  
We study the minimizers
of a criterion that may be viewed as the objective function
when using dropout.  This abstracts away
sampling and optimization issues to focus on the inductive bias, as in
\cite{Bre04,Zha04convex,BJM06,LS10,HL15}.
See Section~\ref{s:prelim} for a complete explanation.

\subsection*{Related work} 
A number of possible explanations have been suggested for dropout's success.
Hinton, et al \cite{HSKSS12} suggest that dropout controls network complexity by restricting the ability to 
co-adapt weights and illustrate how it appears to learn simpler functions at the second layer.
Others \cite{BS13,BAP14} view dropout as an ensemble method combining the different network topologies resulting from the random deletion of nodes.
Wager, et al \cite{WFWL14} observe that in 1-layer networks dropout essentially forces learning on a more challenging distribution
akin to  `altitude training' of athletes.

Most formal analysis of the inductive bias of dropout has
concentrated on the single-layer setting, where a single neuron
combines the (potentially dropped-out) inputs.  Wager, et al \cite{WWL13}
considered
the case that the distribution of label $y$ given feature vector $\bx$
is a member of the exponential family, and the log-loss is used
to evaluate models.  They pointed out that, in this situation, the
criterion optimized by dropout can be decomposed into the original
loss and a term that does not depend on the labels.  
They then gave
approximations to this dropout regularizer and discussed its
relationship with other regularizers.  As we have seen, many aspects
of the behavior of dropout and its relationship to other regularizers
are qualitatively different when there are hidden units.

Wager, et al \cite{WFWL14} considered dropout for learning topics  modeled by a Poisson generative process.
They exploited the conditional independence assumptions of the generative process to
show that the excess risk of dropout training due to training set variation has a term that decays more rapidly than
the straightforward empirical risk minimization, but also has a second additive term related to document length.  
They also discussed situations where the model learned by dropout has small bias.

Baldi and Sadowski \cite{BS14} analyzed dropout in linear networks,
and showed how dropout can be approximated by normalized geometric
means of subnetworks in the nonlinear case.
Gal and Ghahramani \cite{Gal2015DropoutB} described
an interpretation of dropout as an approximation to a deep
Gaussian process.

The impact of dropout (and its relative dropconnect) on
generalization (roughly, how much dropout restricts the search space
of the learner) was studied in \cite{WanEtAl13}. 

In the on-line learning with experts setting, Van Erven, et al \cite{VKW14} showed that applying dropout
in on-line trials leads to algorithms that 
automatically adapt to the input sequence without requiring doubling
or other parameter-tuning techniques.

The rest of the paper is organized as follows.
Section~\ref{s:prelim} introduces our notation and formally defines the dropout model.
We prove that dropout enjoys several scaling invariances that weight-decay doesn't in Section~\ref{s:scale},
and that dropout requires negative weights even in very simple situations in Section~\ref{s:negative.weights}.
Section~\ref{s:dropout.penalty} uncovers various properties of the dropout penalty function.
Section~\ref{s:experiments} describes some simulation experiments.
We provide some concluding remarks in Section~\ref{s:conclusions}.

%% file: prelim.tex
\section{Preliminaries}
\label{s:prelim}

Throughout, we will analyze fully connected layered networks with $K$
inputs, one output, $d$ layers (counting the output, but not the inputs), and $n$ nodes in each hidden layer.
We assume that $n$ is a positive multiple of $K$ and that $K$ is an even 
perfect square and a power of two to avoid unilluminating floor/ceiling clutter in the analysis.
We will call this the \underline{\em{standard
    architecture}}.
We use $\cW$ to denote a particular setting of 
the weights and biases in the network and 
$\cW(\bx)$ to denote the network's output on input $\bx$ using $\cW$.
The hidden nodes are ReLUs, and the output node is linear.
$\cW$ can be decomposed as $(W_1,\bb_1,...,W_{d-1}, \bb_{d-1}, \bw, b)$, where
each $W_j$ is the matrix of weights on connections from the $j-1$st into the $j$th hidden layer,
each $\bb_j$ is the vector of bias inputs into the $j$th hidden layer,
$\bw$ are the weights into the output node, and $b$ is the bias into the output node.

We will refer to a joint probability distribution over examples $(\bx, y)$ 
as an \underline{\em{example distribution}}.
We focus on square loss, so the loss of $\cW$ on example $(\bx, y)$ is $(\cW(\bx) - y)^2$.  
The \underline{\em{risk}}  is the expected loss with respect to an example distribution $P$, 
we denote the risk of $\cW$ as $R_{P}(\cW) \defeq \E_{(\bx, y) \sim P} \left( (\cW(\bx) - y)^2 \right)$.
The subscript will often be omitted when $P$ is clear from the context.

The goal of $L_2$ training 
is to find weights and biases minimizing the 
\underline{\emph{$L_2$ criterion}} with regularization strength $\lambda$:
$
\underline{J_2}(\cW) \defeq  R(\cW) + \frac{\lambda}{2} ||\cW||^2.
$
Here and throughout, we use $||\cW||^2$ to denote the sum of the squares
of the weights of $\cW$.  (As usual, the biases are not
penalized.)
We use \underline{$\cW_{L_2}$} to denote a minimizer of this criterion.
The $L_2$ penalty, $\frac{\lambda}{2} ||\cW||^2$,
is non-negative.  
This is useful, for example, to bound the risk of a minimizer 
$\cW_{L_2}$ of $J_2$, since $R( \cW) \leq J_2 (\cW)$.

Dropout training independently removes nodes in the network.
In our analysis each non-output node is dropped out with the same probability $q$, so $p=1-q$ is the probability that a node is kept.
(The output node is always kept; dropping it out has the effect of cancelling the training iteration.)
When a node is dropped out, the node's output is set to 0. 
To compensate for this reduction, the values of the kept nodes are multiplied by $1/p$.
With this compensation, the dropout can be viewed as injecting zero-mean additive noise at each non-output node \cite{WWL13}.
\footnote{Some authors use a similar adjustment where the weights are 
scaled down
at prediction time instead of inflating the kept nodes' outputs at training time.}

The \underline{\em{dropout process}} is the collection of random choices, for each node in the network, 
of whether the node is kept or dropped out.
A realization of the dropout process is a \underline{\em{dropout pattern}}, 
which is a boolean vector indicating the kept nodes.
For a network ${\cal W}$, an input $\bx$, and dropout
pattern $\cR$, let \underline{$\cD(\cW, \bx, \cR)$} be the output
of $\cW$ when nodes are dropped out or not following $\cR$
(including the $1/p$ rescaling of kept nodes' outputs).
%
The goal of dropout training on an example distribution $P$ is to find weights and biases minimizing the \underline{\emph{dropout criterion}} for a given dropout probability:
\[
J_D (\cW) \defeq \E_{\cR} \E_{(\bx, y) \sim P} \left( (\cD(\cW, \bx, \cR) - y)^2 \right) .
\]
This criterion is equivalent to the expected risk of the dropout-modified network,
and we use $\cW_D$ to denote a minimizer of it.  
Since the selection of dropout pattern and example from $P$ are independent, the order of the two expectations can be swapped, yielding
\begin{equation}
\label{e:swap}
J_D (\cW) = \E_{(\bx, y) \sim P}  \E_{\cR} \left( (\cD(\cW, \bx, \cR) - y)^2 \right).
\end{equation}
Equation~(\ref{e:swap}) is a key property of the dropout criterion.  
It indicates when something is true about the dropout criterion for a family of distributions concentrated on single examples, then
(usually) the same thing will be true for any mixture of these single-example distributions.  

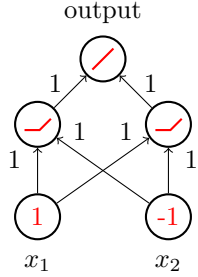
\begin{figure}
\centering
\tikzstyle{neuron}=[draw, circle, thick, black, text=red]
\begin{tikzpicture} [inner sep= 0, minimum size = 17pt, node distance = 35pt]

\node[neuron, label=above:{output}] (E) {\linnode} ;

\node[neuron, below left of =E ] (h1) {
	\relunode
    };
   
\node[ neuron, below right of  = E]     (h2) {
	\relunode
    };
 
\node [ neuron, below of = h1, label=below:{$x_1$}]  (i1) { 1 }; 
\node [ neuron, below of = h2, label=below:{$x_2$}]  (i2) { -1 };

\draw[->] (i1) -- (h1)  node[near end, left] {1};
\draw[->] (i1) -- (h2)  node[near end, above ] {1};
\draw[->] (i2) -- (h1)  node[near end, above ]{1};
\draw[->] (i2) -- (h2)  node[near end, right ] {1};
\draw[->] (h1) -- (E)  node[near end, left] {1};
\draw[->] (h2) -- (E)  node[near end, right] {1};

\end{tikzpicture}    

\caption{A network where the dropout penalty is negative.}
\label{f:example}
\end{figure}

Consider now the example network in Figure~\ref{f:example}.  
The weight parameters $W_1$ and $\bw$ are all-1's, and all of the
biases are 0.
$\cW(1,-1) = 0$ as each hidden node computes 0.
Each dropout pattern indicates the subset of the four lower nodes to be kept, 
and when $q=p=1/2$ each subset is equally likely to be kept.
If $\cR$ is the dropout pattern where input $x_2$ is dropped and the other nodes are kept, then the network computes
$\cD(\cW, (1,-1), \cR) = 8$ (recall that when $p=1/2$ the values of non-dropped out nodes are doubled).
Only three dropout patterns produce a non-zero output, so
if $P$ is concentrated on the example $\bx = (1,-1), y=8$ the dropout criterion is:
\begin{equation}
\label{e:neg.penalty}
J_D(\cW) = \frac{1}{16} (8-8)^2 + \frac{2}{16} (4-8)^2 + \frac{13}{16} (0-8)^2 = 54 .
\end{equation}

As mentioned in the introduction, the \underline{\em{dropout penalty}} of a weight vector for a given example distribution and dropout probability is the amount that the dropout criterion exceeds the risk, $J_D ( \cW ) - R (\cW)$.
Wager, et al \cite{WWL13} show that for 1-layer generalized linear models, the dropout penalty is non-negative.

Since $\cW(1, -1) = 0$, we have $R (\cW) = 64$, and the dropout penalty
is negative in our example.  
This is because the variance in the output due to dropout causes the network to better fit the data (on average) than the network's
non-dropout evaluation. 
In Section~\ref{s:negative.penalty}, we give a necessary
condition for this variance to be beneficial.
As with the dropout criterion, the dropout penalty decomposes into an expectation of penalties over single examples:
\[
J_D ( \cW ) - R (\cW) =
\E_{(\bx, y) \sim P}  \left(  \E_{\cR} \left( (\cD(\cW, \bx, \cR) - y)^2 \right) - (\cW(\bx) - y)^2 \right) .
\]

\begin{definition}
  Define $P_{(\bx,y)}$ as the distribution with half of its weight on example $(\bx,y)$ and half of its weight on example
  $({\bf 0}, 0)$.
\end{definition}

Unless indicated otherwise, we assume  $p = q = 1/2$ for simplicity, although this is not crucial for our results.



%

%% file: scale.tex
\section{Scaling inputs, weights and outputs}
\label{s:scale}

\subsection{Dropout is scale-free}

Here we prove that dropout regularizes deep networks in a 
manner that is independent of the scale of the input features.
In other words, training under dropout regularization 
does not penalize the use of large weights when 
needed to compensate for small input values.

\begin{definition}
For any example distribution $P$, define the 
\underline{\em dropout aversion} with $P$ to be the
maximum, over minimizers $\cW_{\cD}$ of the dropout risk
$J_{\cD} (\cW_{\cD})$, of 
$R_P(\cW_{\cD}) - \inf_{\cW} R_P(\cW)$.
\end{definition}

The dropout aversion of $P$ measures the extent to which
$P$ is incompatible with the inductive bias of dropout, measured by the 
risk gap between the true risk minimizer and the optimizers of the dropout criterion.

\begin{definition}
For example distribution $P$ and square matrix $A$, denote by 
$A \circ P$ the distribution obtained by sampling $(\bx, y)$ from $P$,
and outputting $(A \bx, y)$.
\end{definition}

When $A$ is diagonal and has full rank, then $A \circ P$ is a
rescaling of the inputs, like changing one input from minutes to seconds and another from feet to meters.

\begin{theorem}
\label{t:scale-free}
For any example distribution $P$, and any diagonal full-rank $K \times K$ matrix $A$, the
dropout aversion of $P$ equals the dropout aversion of $A \circ P$.
\end{theorem}
{\bf Proof:}  Choose a network $\cW = (W_1,\bb_1,...,W_{d-1}, \bb_{d-1}, \bw, b)$.  
Let $\cW' = (W_1 A^{-1},\bb_1...,W_{d-1},\bb_{d-1}, \bw,b)$.  For any $\bx$,
$\cW(\bx) = \cW'( A \bx)$, as $A^{-1}$ undoes the effect of $A$ before
it gets to the rest of the network, which is unchanged. 
Furthermore, for any dropout
pattern $\cR$, we have $\cD(\cW, \bx, \cR) = \cD(\cW', A \bx, \cR)$.
Once again $A^{-1}$ undoes the effects of $A$ on kept nodes (since $A$ is diagonal), 
and the rest of the network $\cW'$ is modified by $\cR$ an a manner paralleling
$\cW$.  
Thus, there is bijection between networks $\cW$ and networks
$\cW'$ with $J_{\cD}(\cW') = J_{\cD}(\cW)$ and $R(\cW') = R(\cW)$, 
yielding the theorem.
\qed

Theorem~\ref{t:scale-free} indicates that some common normalizations of the input features (e.g. to have unit variance) do not affect the 
quality
of the dropout criterion minimizers,  but normalization might change  the speed of convergence and which minimizer is reached.
Centering the features has slightly different properties.  
Although it is easy to use the biases to define a  $\cW'$ that ``undoes'' the centering in the non-dropout computation,  different $\cW'$  
appear to be
required for different dropout patterns, breaking the bijection exploited in Theorem~\ref{t:scale-free}.

As we will see in 
Section~\ref{s:weight_decay_scale}, 
weight decay does not enjoy such scale-free status.  

%
%

\subsection{Dropout's invariance to parameter scaling}

Next, we describe an equivalence relation among parameterizations for dropout networks of depth $d\geq 2$. 
Basically, scaling the parameters at a level creates a corresponding scaling of the output.
(A similar observation was made in a somewhat different context
by Neyshabur, et al \cite{NTS15}.)
\begin{theorem}
\label{t:scale-unscale}
For any input $\bx$, dropout pattern $\cR$, any network
$\cW = (W_1,\bb_1,...,W_{d-1}, \bb_{d-1}, \bw, b)$, 
and any positive $c_1,..., c_d$,
if 
\begin{equation}
\label{e:cWprime.def}
\cW' = \left(c_1 W_1,c_1 \bb_1,
            c_2 W_2,c_1 c_2 \bb_2,
...,c_{d-1} W_{d-1}, \left( \prod_{j=1}^{d-1} c_j \right) \bb_{d-1}, 
        c_d \bw, \left( \prod_{j=1}^{d} c_j \right) b\right),
\end{equation}
then $\cD( \cW', \bx, \cR) = \left(\prod_{j=1}^d c_j \right) \cD( \cW, \bx, \cR)$.
In particular, if  $\prod_{j=1}^d c_j = 1$,
then for any example distribution $P$, networks $\cW$ and $\cW'$ have the same dropout criterion, dropout penalty, and
expected loss.
\end{theorem}

Note that the re-scaling of the biases at
layer $j$ depends not only on the rescaling of the connection weights at layer $j$, but also the re-scalings at lower layers.

{\bf Proof:}  Choose an input $\bx$ and a dropout pattern $\cR$.
Define $\cW'$ as in (\ref{e:cWprime.def}).
For each hidden layer $j$, let 
\renewcommand{\th}{\tilde{h}}
$(h_{j1},...,h_{jn})$ be the $j$th hidden layer when
applying $\cW$ to $\bx$ with $\cR$, and let
$(\th_{j1},...,\th_{jn})$ be the $j$th hidden layer when
applying $\cW'$ instead.  
By induction, for all $i$, 
$\th_{ji} = \left( \prod_{\ell \leq j} c_\ell \right) h_{ji}$; the key
step is that the pre-rectified value used to compute
$\th_{ji}$ has the same sign as for $h_{ji}$, since rescaling by 
$c_j$ preserves the sign.  Thus the same units are 
zeroed out
in
$\cW$ and $\cW'$ and
$\cD(\cW, \bx, \cR) = (\prod_j c_j) \cD(\cW', \bx, \cR)$.
When $\prod_j c_j = 1$, this implies
$\cD(\cW, \bx, \cR) = \cD(\cW', \bx, \cR)$.
Since this is true for all $\bx$ and
$\cR$, we have $J_{\cD}(\cW) = J_{\cD}(\cW')$.
Since, similarly, $\cW(\bx) = \cW'(\bx)$ for all $\bx$, so
$R(\cW) = R(\cW')$, the dropout penalities for $\cW$ and
$\cW'$ are also the same.
\qed

Theorem~\ref{t:scale-unscale} implies that 
the dropout criterion never has isolated minimizers,
since one can continuously up-scale the weights on one layer with a compensating down-scaling at another
layer to get a contiguous family of networks computing the same function and having the same dropout criterion.
It may be possible to exploit the parameterization equivalence of Theorem~\ref{t:scale-unscale} in training by
using canonical forms for the equivalent networks or switching to an equivalent networks whose gradients have better properties.
We leave this question for future work.

\subsection{Output scaling with dropout}

Scaling the output values of an example distribution $P$  does affect
the aversion, but in a very simple and natural way.
\begin{theorem}
\label{t:output.scaling}
For any example distribution $P$, if $P'$ is obtained from $P$
by scaling the outputs of $P$ by a positive constant $c$, the
dropout aversion of $P'$ is $c^2$ times the dropout aversion of $P$.
\end{theorem}
{\bf Proof:} 
%
If a network $\cW $ minimizes the dropout
criterion for $P$, then the network $\cW'$ obtained by
scaling up the weights and bias for the output unit by $c$
minimizes 
the dropout criterion for $P'$, and
for any $\bx, y$, and dropout pattern $\cR$, 
$ ( \cD(\cW, \bx, \cR) - y)^2 =  ( \cD(\cW', \bx, \cR) - cy)^2 / c^2$.
\qed


\subsection{Scaling properties of weight decay}
\label{s:weight_decay_scale}

Weight decay does not have the same scaling properties as dropout.
Define the \emph{weight-decay aversion} analogously to the dropout aversion.


We analyze the $L_2$ criterion for depth-2 networks
in Appendix~\ref{s:not-scale-free}, resulting in the following theorem.
%
Our proof shows that, despite the non-convexity of the $L_2$ criterion,
it is still possible to identify a closed form for one of its optimizers.

%

\begin{theorem}
\label{t:not-scale-free}
Choose an arbitrary number $n$ of hidden nodes, and $\lambda > 0$. 

The weight-decay aversion of $P_{ ( \bx, 1 ) }$ is $\min ( \frac{1}{4},
\frac{\lambda^2}{ \bx \cdot \bx })$.
\end{theorem}

Theorem~\ref{t:not-scale-free} shows that, unlike dropout, the weight
decay aversion \emph{does} depend on the scaling of the input features.
%

Furthermore, when  $2 \lambda > \sqrt{ \bx \cdot \bx }$, the weight-decay criterion for $P_{ (\bx,1) }$ has only a single isolated optimum
weight setting%
\footnote{Since the output node puts weight 0 on each hidden node and the biases are unregularized, this optimum actually represents a class of networks differing only in the irrelevant biases at the hidden nodes.  One can easily construct other cases when weight-decay has isolated minima in this sense,
for example when $n=2$ and there is equal probability on $\bx$ and $-\bx$, both with label 1.}
 -- all weights set to zero and bias 1/2 at the output node. 
This means that  weight-decay in 2-layer networks can completely regularize away significant signal in the sample \emph{even when $\lambda$ is finite},
contrasting starkly with weight-decay's behavior  in 1-layer networks.


The ``vertical'' flexibility to rescale weights between layers enjoyed by dropout (Theorem~\ref{t:scale-unscale})
does not hold for $L_2$: one can always drive the $L_2$ penalty to infinity by scaling
one layer up by a large enough positive $c$, even while scaling another
down by $c$.
On the other hand, the proof of Theorem~\ref{t:not-scale-free} 
shows that the $L_2$ criterion has an alternative ``horizontal'' flexibility involving the rescaling of weights across nodes on the
hidden layer (under the theorem's assumptions).
Lemma~\ref{l:activation.by.weight} shows that at the optimizers each
hidden node's contributions to the output are a constant (depending on the input) times their contribution to the
the $L_2$ penalty.  
Shifting the magnitudes of these contributions between hidden nodes leads to alternative weights that
compute the same value and have the same weight decay penalty.  
This is a more general observation than the permutation symmetry between hidden nodes because any portion of a hidden node's contribution can be shifted to another hidden node.

%
%
%

%% file: negative_weights.tex
\section{Negative weights for monotone functions}
\label{s:negative.weights}

\newcommand{\cWn}{\cW_{\mathrm{neg}}}

If the weights of a unit are non-negative, then the unit computes a
monotone function, in the sense that increasing any input while
keeping the others fixed increases the output.  
The bias does not affect a node's monotonicity.
A network of monotone units is also monotone.  

We first present our theoretical results for many features (Section~\ref{s:basic}) and few features (Section~\ref{s:few}), and then discuss the implication
of these results in Section~\ref{s:general}.

\subsection{The basic case -- many features}
\label{s:basic}

In this section, we analyze the simple distribution $P_{({\bf 1}, 1)}$ that assigns
probability $1/2$ to the example $(0,...,0), 0$, and probability $1/2$
to the example $(1,...,1), 1$.  This is arguably the simplest monotone
function.  Nevertheless, we prove that dropout 
uses
negative weights to fit this data.  

The key 
intuition is that optimizing the dropout criterion
requires controlling the variance.  
Negative weights at the hidden nodes can be used to control the variance
due to dropout at the input layer.
When there are enough hidden nodes this becomes so beneficial that every minimizer of the dropout criterion
uses such negative weights.

\begin{theorem}
\label{t:negative_weights}
For the standard architecture, if $K > 18$ and 
$n$ is large enough relative to $K$ and $d$,
every 
optimizer of the dropout criterion for $P_{({\bf 1}, 1)}$ uses at
least one negative weight.
\end{theorem}

To prove Theorem~\ref{t:negative_weights},
we first calculate $J_{\cD}(\cWn)$ for a network $\cWn$ that
uses negative weights, and then prove a lower bound greater than this value that holds for
all networks using only non-negative weights.

All of the biases in $\cWn$ are $0$.

A key building block in the definition of $\cWn$ is a block of 
hidden units that we call the \underline{\em{first-one gadget}}.
Each such block has $K$ hidden nodes, and 
takes its input from the $K$ input nodes.  The $i$th hidden node in
the block takes the value $1$ if the $K$th input node is $1$, and
all inputs $x_{i'}$ for $i' < i$ are $0$.  This can be accomplished with
a weight vector $\bw$ with $w_{i'} = -1$ for $i' < i$, with
$w_{i} = 1$, and with $w_{i'} = 0$ for $i' > i$.  
The first hidden layer of $\cWn$ comprises $n/K$ copies of the
first-one gadget.

Informally, this construction removes most of the variance in the number 
of $1's$ in the input, as recorded in the following lemma.
\begin{lemma}
On any input $\bx \in \{0,1 \}^n$ except $(0,0,...,0)$, the
sum of the values on the first hidden layer of $\cWn$ is
exactly $n/K$.
\end{lemma}

The weights into the remaining hidden layers of $\cWn$ are all
$1$, and all the weights into the output layer take a value 
$c \defeq \frac{K^2}
           {2 n^{d-1} \left(1 + \frac{K}{n} \right) 
              \left(1 + \frac{1}{n} \right)^{d-2} }$
, chosen to minimize the dropout criterion for the network.
The following lemma analyzes $\cWn$.
\begin{lemma}
\label{l:neg.upper}
$J_{\cD} (\cWn) = 
    \frac{1}{2} \left( 1 - \frac{(1 - 2^{-K})}{\left(1 + \frac{K}{n} \right) 
       \left(1 + \frac{1}{n} \right)^{d-2} } \right).$
\end{lemma}

When $n$ is
large relative to $K$ and $d$, the $\left(1 + \frac{K}{n} \right)  \left(1 + \frac{1}{n} \right)^{d-2}$ denominator in Lemma~\ref{l:neg.upper}
approaches 1, so $J_{\cD} ( \cWn )$ approaches $2^{-K} / 2$ in this case.
Lemma~\ref{l:pos.lower} below gives a larger lower bound for any network
with all positive weights.
In the concrete case when $d=2$ and $n=K^3$, then Lemma~\ref{l:neg.upper} implies $J_{\cD} ( \cWn ) < 1/K^2 $.

\medskip

{\bf Proof } (of Lemma~\ref{l:neg.upper}):
Consider a computation of
$\cWn(1,1,...,1)$ under dropout and let $\hy$ be the (random) output.  
Let $k_0$ be the number of input nodes
kept, and, for each $j \geq 2$, let $k_j$ be the number of nodes in the $j$th hidden layer kept.
Call the node in each first-one gadget that computes 1 a \emph{key} node, and
if no node in the gadget computes 1 because the input is all dropped, 
arbitrarily make the gadget's first hidden node the key node.
This ensures there is exactly one key node per gadget, and every non-key node computes 0.
Let $k_1$ be the number of kept key nodes on the first hidden layer.
If $k_0 = 0$,
the output $\hy$ of the network is $0$.  
Otherwise, $\hy = c 2^d \prod_{j=1}^{d-1} k_j$.

Note that $k_0$ is zero with probability $2^{-K}$.  Whenever 
$k_0 \geq 1$, $k_1$ is distributed as $B(n/K,1/2)$.
Each other $k_j$ is distributed as $B(n, 1/2)$, and 
$k_1, k_2,...,k_{d-1}$ are independent of one another.
\begin{align*}
& \E(\hy) 
 = \Pr(k_0 \geq 1) c 2^{d} \E[k_1 | k_0 \geq 1] \prod_{j=2}^{d-1} \E[k_j]
 = (1 - 2^{-K}) c 2^{d}  \left(\frac{n}{2K} \right) \left(\frac{n}{2}\right)^{d-2}
 = \frac{2 c}{K} (1 - 2^{-K}) n^{d-1}. \\
\end{align*}
Using the value of the second moment of the binomial, 
we get
\begin{align*}
& \E(\hy^2) 
  =  
  \E\left[ \left( \ind_{k_0 \geq 1} c 2^d \prod_{j=1}^{d-1} k_j \right)^2\right] 
  =  
  4 c^2 (1 - 2^{-K})  
   \left(\frac{n}{K} \right)  \left(\frac{n}{K} + 1 \right) 
    n^{d-2} (n+1)^{d-2} \\
& =  
  \frac{4 c^2 (1 - 2^{-K})}{K^2}
    n^{2(d-1)}
    \left(1 + \frac{K}{n} \right) 
    \left(1 + \frac{1}{n} \right)^{d-2}.
\end{align*}
Thus, 
\begin{align*}
J_{\cal D}(\cWn) 
	&=  \frac{1}{2} 
      \left( 1  - 2 \E(\hat y) + \E( \hat y^2 ) \right) \\
	& = \frac{1}{2}
            \left(
            1  - \frac{4 c}{K} (1 - 2^{-K}) n^{d-1}
            +   \frac{4 c^2 (1 - 2^{-K})}{K^2} n^{2(d-1)}
      \left(1 + \frac{K}{n} \right) 
    \left(1 + \frac{1}{n} \right)^{d-2} \right) \\
 & = \frac{1}{2} \left( 1 - \frac{(1 - 2^{-K})}{\left(1 + \frac{K}{n} \right) 
              \left(1 + \frac{1}{n} \right)^{d-2}}
               \right),
\end{align*}
since $c = \frac{ K }
                        {2 n^{d-1} \left(1 + \frac{K}{n} \right) 
                         \left(1 + \frac{1}{n} \right)^{d-2}}$,
completing the proof.\qed

Next we prove a lower bound on $J_{\cD}$ for networks with nonnegative weights.
Let $\cW$ be an arbitrary such network.

Our lower bound will use a property of the function computed by $\cW$ that we now define.
\begin{definition}

A function $\phi : \R^K \rightarrow \R$ is \underline{\em supermodular} if
for all $\bx, \bdelta_1, \bdelta_2 \in  \R^K$ where 
$\bdelta_1, \bdelta_2 \geq \mathbf{0}$
\[
\phi(\bx) + \phi(\bx + \bdelta_1 + \bdelta_2 ) \geq \phi(\bx + \bdelta_1) + \phi(\bx+\bdelta_2),
\]
or equivalently:
\[
\phi(\bx + \bdelta_1 + \bdelta_2 ) - \phi(\bx+\bdelta_2)
\geq \phi(\bx + \bdelta_1) - \phi(\bx)
\]
\end{definition}
The latter indicates that adding $\bdelta_1$ to the bigger input $\bx+\bdelta_2$ has at least as large an effect as
adding it to the smaller input $\bx$.

Since $\cW$ has all non-negative weights, it computes a supermodular function of its inputs.
(This fact may be of independent interest.)
\begin{lemma}
 \label{l:supermod.overall}
If a network has non-negative weights and its activation functions
$\sigma(\cdot)$ are convex, continuous, non-decreasing, and
differentiable except on a finite set, then the network computes a
supermodular function of its input $\bx$.
\end{lemma}
{\bf Proof:}  We will prove by induction over the layers that, for any unit
$h$ in the network, if $h(\bx)$ is the output of unit $h$ when
$\bx$ is the input to $\cW$, then $h(\cdot)$ is a supermodular function of
its input.

The base case holds since each input node $h$ outputs the corresponding component of the input,  
and  $(\bx+\bdelta_1)  - \bx  = (\bx+ \bdelta_1 + \bdelta_2) - (\bx + \bdelta_2)$.

Now, for the inductive step, let $\bw$ be the weight vector for node
$h$, let $b$ be its bias, and $\sigma(\cdot)$ be its activation function.
Let $I(\bx)$, $I(\bx+\bdelta_1)$, $I(\bx+\bdelta_2)$, and $I(\bx+\bdelta_1+\bdelta_2)$ be the inputs to node $h$ 
when the inputs to
the network are $\bx$, $\bx+\bdelta_1$, $\bx+\bdelta_2$ and $\bx+\bdelta_1+\bdelta_2$
respectively.  

By induction, these inputs to node $h$ (componentwise) satisfy 
\[
I(\bx + \bdelta_1) - I(\bx) \leq I(\bx + \bdelta_1 + \bdelta_2) - I(\bx+\bdelta_2).
\]
Therefore, since $\bw$, $\bdelta_1$, and $\bdelta_2$ are non-negative, 
the interval $[ \bw \cdot I(\bx+ \bdelta_2)+b, \bw \cdot I(\bx + \bdelta_1 + \bdelta_2) +b ] $
is at least as long and starts at least as high as the interval
$[ \bw \cdot I(\bx)+b, \bw \cdot I(\bx + \bdelta_1)+b ] $.
Since $\sigma$ is continuous and differentiable except on a finite
set, we have 
\begin{align*}
h( I( \bx+\bdelta_1 ) ) - h( I(\bx) ) 
	&=    \int_{\bw \cdot  I(\bx) +b }^{\bw \cdot I(\bx + \bdelta_1) +b } \sigma'(z) dz \\
	& \leq \int_{\bw \cdot  I(\bx+\bdelta_2) +b }^{\bw \cdot I(\bx + \bdelta_1 + \bdelta_2) +b } \sigma'(z) dz \;\;\; 
       \mbox{(since $\sigma'$ is non-decreasing) }\\
	& = h( I(\bx + \bdelta_1 + \bdelta_2) ) - h( I(\bx+\bdelta_2) ).
\end{align*}
\qed

\begin{definition}
Let $\br_0 \in \{0,1\}^K$ be the dropout pattern concerning the
  input layer, and let $\cR'$ be the dropout
  pattern concerning the rest of the network, so that the dropout pattern
  $\cR = (\br_0, \cR')$.

For each $\ell \in \{ 0,...,K\}$, let $\psi_{\cW} (\ell)$ be
the average output of $\cW$ under dropout when $\ell$ of the
inputs are kept: i.e.,
\[
\psi_{\cW} (\ell) 
  = \E \left( \cD(\cW, 1^K, (\br_0, \cR'))  \middle\vert  \sum_j r_{0j} = \ell \right).
\]
\end{definition}

\begin{lemma}
 \label{l:convex.count}
For any $\ell \in \{ 1,..., K - 1, \}$,
\[
\psi_{\cW} (\ell + 1) - \psi_{\cW} (\ell)
  \geq \psi_{\cW} (\ell) - \psi_{\cW} (\ell - 1) .
\]
\end{lemma}
{\bf Proof:}  Generate $\bu$, $i$ and $j$ randomly by,
first choosing $\bu$ uniformly at random from among
bit vectors with $\ell$ ones, then choosing
$i$ uniformly from the $0$-components of $\bu$, and $j$ 
uniformly from the $1$-components of $\bu$.  By
Lemma~\ref{l:supermod.overall}, 
\begin{equation}
\label{e:supermod.onestep}
\cW(\bu + \be_i) - \cW(\bu) \geq \cW(\bu - \be_j + \be_i) - \cW(\bu - \be_j)
\end{equation}
always holds.  Furthermore, $\bu + \be_i$ is uniformly distributed among
bit vectors with $\ell + 1$ ones, $\bu - \be_j$ is uniformly distributed among
bit vectors with $\ell - 1$ ones, and $\bu + \be_i - \be_j$ is
uniformly distributed among bit vectors with $\ell$ ones.
This is true for $\cW$, but it is also true for
any network obtained by dropping out some of the hidden nodes of
$\cW$.  Thus
\begin{align*}
& \psi_{\cW} (\ell + 1) - \psi_{\cW} (\ell) \\
& = 
\E (\cD(\cW, 1^K, (\br_0, \cR') | \sum_j r_{0j} = \ell+1)) 
- \E (\cD(\cW, 1^K, (\br_0, \cR') | \sum_j r_{0j} = \ell))  \\
& = 
\E (\cD(\cW, \br_0, (1^K, \cR') | \sum_j r_{0j} = \ell+1)) 
- \E (\cD(\cW, \br_0, (1^K, \cR') | \sum_j r_{0j} = \ell))  \\
& \geq 
\E (\cD(\cW, \br_0, (1^K, \cR') | \sum_j r_{0j} = \ell)) 
- \E (\cD(\cW, \br_0, (1^K, \cR') | \sum_j r_{0j} = \ell-1)) 
  \;\;\;\mbox{(by (\ref{e:supermod.onestep}))} \\
& = \psi_{\cW} (\ell) - \psi_{\cW} (\ell-1),
\end{align*}
completing the proof. \qed

We will use the following lower bound on the tail of the binomial.
(Many similar lower bounds are known.)
\begin{lemma}
\label{l:binomial.lower}
If $X$ is distributed according to $\Bin(n,1/2)$, then 
\[
\Pr(X < n/2 - \sqrt{n}/4) = \Pr(X > n/2 + \sqrt{n}/4) \geq 1/4.
\]
\end{lemma}
{\bf Proof:} 
Using the fact that, for any $i$, $\Pr(X = i) \leq 1/\sqrt{n}$, we get 
$\Pr(| X - n/2 | < \sqrt{n}/4) \leq 1/2$, so $\Pr(X < n/2 - \sqrt{n}/4) \geq 1/4$.  \qed

Now we are ready for the lower bound on $J_{\cD} (\cW).$
\begin{lemma}
\label{l:pos.lower}
If $K > 18$ and the weights in $\cW$ are non-negative, then
$J_{\cD} (\cW) \geq \frac{1}{36 K}$.
\end{lemma}
{\bf Proof:}  
Assume to the contrary that $J_{\cD} (\cW) < \frac{1}{36 K}$.
First, note that 
$\psi_{\cW}(0) \leq \sqrt{\frac{1}{18 K}}$, 
or else the contribution to $J_{\cD} (\cW)$ due to the $(0,0),0$ example is at least $\frac{1}{36 K}$.
Applying Lemma~\ref{l:binomial.lower}, we have
\begin{equation}
\label{e:shoulders}
\psi_{\cW}(K/2 - \sqrt{K}/4) >  1- \sqrt{\frac{2}{9 K}} \mbox{ and } \psi_{\cW}(K/2 + \sqrt{K}/4 ) < 1+ \sqrt{\frac{2}{9 K}}
\end{equation}
as otherwise the contribution of one of the tails to $J_{\cD} (\cW)$ will be at least 
$\frac{1}{36 K}$ for the $(1, \ldots,1), 1$ example.
We will contradict this small variation of $\psi_{\cW}( \ell )$ around $K/2$.
The bounds on $\psi_{\cW}(0)$
and $\psi(K/2  - \sqrt{K}/4)$ and
Lemma~\ref{l:convex.count} imply  that $\psi_{\cW}( \ell )$ grows rapidly when $\ell$ is around $K/2$, in particular:
\[
\psi(K/2 - \sqrt{K}/4 + 1)  - \psi(K/2 - \sqrt{K}/4) 
	\geq   \frac{1 - \sqrt{\frac{2}{9K}} - \sqrt{\frac{1}{18 K}} }{ K/2- \sqrt{K}/4 }  
	> \frac{1}{\sqrt{9/32} K},
\]
since $K > 18$.
Now using Lemma~\ref{l:convex.count} repeatedly shows that
\begin{align*} 
\psi(K/2 + \sqrt{K}/4)  - \psi(K/2 - \sqrt{K}/4) 
	& >  \frac{\sqrt{K}}{2} \times \frac{1}{ \sqrt{9/32} K} = 2 \sqrt{\frac{2}{9 K}},
\end{align*}
which contradicts (\ref{e:shoulders}), completing the proof.
\qed


Putting together Lemmas~\ref{l:neg.upper} and \ref{l:pos.lower}
immediately proves Theorem~\ref{t:negative_weights}, since for 
$K > 18$ and 
large enough $n$, the criterion for $\cWn$ must be less than
the criterion for any network with all non-negative weights.

\subsection{The case when $K=2$}
\label{s:few}

Theorem~\ref{t:negative_weights} uses the assumption that $K > 18$ and $n$ is large
enough;  is the lower bound on $K$ really necessary?  Here we show that it is not, by
treating the case that $K=2$.  
\begin{theorem}
\label{t:negative_weights.K=2}
For the standard architecture, if $K = 2$, for any fixed $d$ and large
enough $n$, every optimizer of the dropout criterion for $P_{(1,1),1}$ uses
negative weights.
\end{theorem}
{\bf Proof:}  
Define $\cWn$ as in the proof of Lemma~\ref{l:neg.upper}, except that
the output layer has a bias of $1/5$.

We claim that
\begin{equation}
\label{e:neg.K=2}
\lim_{n \rightarrow \infty} J_{\cD} (\cWn) = 1/10 .
\end{equation}
To see this, consider the joint input/label distribution under dropout:
\begin{align*}
& \Pr((0,0),0) = 1/2 \\
& \Pr((0,0),1) = 1/8 \\
& \Pr((2,0),1) = 1/8 \\
& \Pr((0,2),1) = 1/8 \\
& \Pr((2,2),1) = 1/8.
\end{align*}

Due to the bias of $1/5$ on the output, $\cWn(0,0) = 1/5$.  Thus,
contribution to $J_{\cD}$ from examples with $\bx = (0,0)$ in this
joint distribution is $1/2 \times (1/5)^2 + 1/8 \times (4/5)^2 =
1/10$.

Now, choose $\bx \neq (0,0)$.  If, after dropout, the input is $\bx$,
each node in the hidden layer closest to the input computes $1$.  
Arguing exactly as in the proof of Lemma~\ref{l:neg.upper}, in
such cases, 
\[
\E((\hy - 1)^2)
     = 1 - \frac{1}{\left(1 + \frac{2}{n} \right) 
              \left(1 + \frac{1}{n} \right)^{d-2}}.
\]
This proves (\ref{e:neg.K=2}).  

Now, let $\cW$ be an arbitrary network with non-negative weights.

For our distribution,
\[
J_{\cal D}(\cW)
=  \frac{ \E_\cR \left( ( \cD(\cW, (0,0), \cR) - 0)^2  \right) + 
		\E_\cR \left( ( \cD(\cW, (1,1), \cR) - 1)^2  \right)}
		{2}  .
\]


Let $V_{00} = \E(\cW(0,0)) , V_{22} = \E(\cW(2,2)), V_{20} = \E(\cW(2,0)), V_{02} = \E(\cW(0,2))$
where the expectations are taken with respect to the dropout patterns at the hidden nodes 
(with no dropout at the inputs).
Since each dropout pattern over the hidden nodes defines a particular network, 
and Lemma~\ref{l:supermod.overall} holds for all of them, the relationships also hold for the expectations, so
\[
V_{22} \geq V_{20} + V_{02} - V_{0}.
\]
Using this $V$ notation, handling the dropout at the input explicitly, and the bias-variance decomposition 
keeping just the bias terms we get:

\begin{align}
J_{\cal D}(\cW) & \geq  \frac{( V_{00} - 0 )^2   + 
		\left( ( V_{00} - 1)^2  + (V_{22} - 1)^2 + (V_{20} - 1)^2 + (V_{02} - 1)^2  \right) / 4 } { 2 }   \label{e:Vapprox} \\
8 J_{\cal D}(\cW) & \geq 4 (V_{00}-0)^2 + (V_{00}-1)^2 + (V_{22} - 1)^2 + (V_{20} - 1)^2 + (V_{02} - 1)^2.
\end{align}

We will continue lower bounding the RHS.
We can re-write $V_{22}$ as $V_{20} + V_{02} - V_{00} + \epsilon$ where $\epsilon \geq 0$.
This is convex and symmetric in $V_{02}$ and $V_{20}$ so they both take the same value
at the minimizer of the RHS, so we proceed using $V_{20}$ for this common minimizing value.

\begin{align*}
8 J_{\cal D}(\cW) 
 & \geq ( 2 V_{20} -  V_{00} +  \epsilon - 1)^2 + 2 ( V_{20} - 1)^2 + ( V_{00} - 1)^2 + 4  V_{00}^2.
\end{align*}

Differentiating with respect to $V_{20}$, we see that the RHS is minimized when $V_{20} = (2 + V_{00} - \eps) / 3$,
giving
\[
8 J_{\cal D}(\cW) 
  \geq   \frac{(  V_{00} - \eps- 1)^2 }{ 3 } + (V_{00} - 1)^2 + 4  V_{00}^2.
\]
 
If  $V_{00} \geq 1$, then $J_{\cal D} \geq 1/2$ (just from the (0,0),0 example) and when $V_{00} < 1$ 
the RHS of the above is minimized for non-negative $\eps$ when $\eps=0$.  
Using this substitution, the minimizing value of $V_{00}$ is 1/4 giving
\begin{align*}
8 J_{\cal D}(\cW)  & \geq 1 \\
J_{\cal D}(\cW)  & \geq 1/8.
\end{align*}

Combining this with (\ref{e:neg.K=2}) completes the proof. \qed

\subsection{More general distributions and implications}
\label{s:general}

In the previous sub-section we analyzed the distribution $P$ 
over the 2-feature examples $(1,1), 1$ and $(0,0), 0$.
However, these two examples can be embedded in a 
larger feature space by using any fixed vector of additional feature values, creating, for instance,
a distribution over $(\underline{0, 1}, 0, 0, \underline{0, 1/2, 1}), 0$ and $(\underline{0,1}, 1,1, \underline{0 , 1/2, 1}), 1$ 
(with the additional features underlined)
The results of Section~\ref{s:few} still apply to distribution over these extended examples after
defining $\cWn$ network to have zero weight on the additional features, and noticing that any weight on the additional features 
in the positive-weight network $\cW$ can be simulated using the biases at the hidden nodes.

It is particularly interesting when the additional features all take the value 0, we call these \underline{\emph{zero-embeddings}}.  
Every network $\cW$ with non-negative weights has
$J_{\cal D} (\cW)  \geq 1/8$ on each of these \emph{zero-embeddings} of $P$.
On the other hand, a single $\cWn$ network with $n/K$ copies of the $K$-input first-one gadget
has $J_{\cal D} (\cWn) \approx 1/10$ simultaneously for all of these zero-embeddings of $P$
(when $n >> K$).  

Any source distribution over $\{0,1\}^K \times \{0,1\}$ that puts probability 1/2 on the 
the $\mathbf{0}, 0$ example and distributes the other $1/2$ probability over examples where exactly two inputs are one
is a mixture of zero-embeddings of $P$, and thus 
$J_{\cal D} (\cW)  \geq 1/8$ while 
$J_{\cal D} (\cWn) \approx 1/10$ for this mixture
and optimizing the dropout criterion requires negative weights.

In our analysis the negative weights used by dropout are counterintuitive for fitting monotone behavior, but are needed to control the 
variance due to dropout.  
This suggests that dropout may be less effective when layers with sparse activation patterns are fed into wider layers,
as dropout training can hijack 
part of
the expressiveness of the wide layer to control the artificial variance due to dropout 
rather than fitting the underlying patterns in the data.

%

%% file: growth.tex
\subsection{Growth of the dropout penalty as a function of $d$}
\label{s:growth}

Weight-decay penalizes large weights, while Theorem~\ref{t:scale-unscale} shows that
compensating rescaling of the weights does not affect the dropout penalty or criterion.
On the other hand, dropout can be more sensitive to the calculation of large outputs than weight decay,
and large outputs can be produced in deep networks using only small weights.
We make this observation concrete by exhibiting a family of networks where 
the depth and desired output are linked while the size of individual weights remains constant.
For this family, the dropout penalty grows exponentially in the depth $d$ (as opposed to linearly for weight-decay), 
suggesting that dropout training is less willing to fit the data in this kind of situation.

\begin{theorem}
\label{t:growth}
If $\bx = (1,1,...,1)$ and $0 \leq y \leq K n^{d-1}$,
for $P_{\bx, y}$
there are weights $\cW$ for the standard architecture with 
$R(\cW) = 0$ such that (a) every weight has magnitude
at most one, but (b) $J_{\cD}(\cW) \geq \frac{y^2}{K+1}$, whereas (c) $J_2(\cW) \leq \frac{\lambda y^{2/d}}{2}(Kn + n^2 (d-2) + n)$.
\end{theorem}
{\bf Proof:}  Let $\cW$ be the network whose weights are all 
$c = \frac{y^{1/d}}{K^{1/d} n^{(d-1)/d}}$ and biases are all $0$,
so that the $L_2$ penalty is the number of weights times $\lambda c^2/2$.
It is a simple induction to show that, for these weights and input
$(1,1,...,1)$, 
the value computed at each hidden node on level $j$ is $c^j K n^{j-1}$, so the 
the network outputs $c^d K n^{d-1}$, and has zero square loss 
(since ${\cW}(\bx) = c^d K n^{d-1} =  y$).

Consider now dropout on this network.
This is equivalent to changing all of the weights from $c$ to $2 c$
and, independently with probability $1/2$, replacing the 
value
of each node with 0. 
For a fixed dropout pattern, each node on a given layer has the same
weights, and receives the same (kept) inputs.  Thus, the value
computed at every node on the same layer is the same.
For each $j$, let $H_j$ be the value computed by the units in the
$j$th hidden layer.  

If $k_0$ is the number of input nodes kept under dropout, and,
for each $j \in \{ 1,..., d-1 \}$, $k_{j}$ is the number of 
hidden nodes kept in layer $j$, a straightforward induction
shows that, for all $\ell$, we have 
$H_{\ell} = (2 c)^{\ell} \prod_{j = 0}^{\ell-1} k_j$, so that the output
$\hy$ of the network is $(2 c)^{d} \prod_{j = 0}^{d-1} k_j$.

Using a bias-variance decomposition, 
$
\E((\hy - y)^2) = (\E[\hy] - y)^2 + \Var(\hy).
$
Since each $k_j$ is binomially distributed, 
and $k_0,...,k_{d-1}$ are independent, we have
$
\E(\hy) = (2c)^d (K/2) (n/2)^{d-1} = c^d K n^{d-1} = y,
$
so
$
\E((\hy - y)^2) = \Var(\hy).
$
Since
$
\E(\hy^2) = (2c)^{2d} (K (K+1)/4) (n (n+1)/4)^{d-1} 
          = y^2 (1 + 1/K) (1 + 1/n)^{d-1},
$
we have
$\Var(\hy)
 = \E(\hy^2) - \E(\hy)^2
 =  y^2 ((1 + 1/K) (1 + 1/n)^{d-1} - 1)
 \geq y^2/K,
$
completing the proof.
\qed

If $y = \exp(\Theta(d))$, the dropout penalty grows exponentially in
$d$, whereas the $L_2$ penalty grows polynomially.

%% file: negative_penalty.tex
\subsection{A necessary condition for negative dropout penalty}
\label{s:negative.penalty}

Section~\ref{s:prelim} contains an example where the dropout penalty is
negative.  The following theorem provides a necessary condition.
\begin{theorem}
\label{t:negative.penalty}
The dropout penalty can be negative. 
For all example distributions, a necessary condition for this in rectified linear networks is that either a weight, input, or bias is negative.
\end{theorem}
{\bf Proof:}
\cite{BS14} 
show that for networks of linear units (as opposed to the non-linear rectified linear units we focus on) the network's output without dropout equals the expected output over dropout patterns, so in our notation:
$\cW (\bx)$ equals $\E_\cR (\cD(\cW, \bx, \cR) ) .$
Assume for the moment that the network consists of linear units and the 
example distribution is concentrated on the single example $(\bx,y)$.
Using the bias-variance decomposition for square loss and this property of linear networks,
\[
J_D (\cW) \!=\! \E_{\cR} \left( (\cD(\cW, \bx, \cR) \!-\! y)^2 \right)  
	\!=\!  (\E_\cR (\cD (\cW, \bx, \cR) \!-\! y)^2 + \Var_\cR ( \cD (\cW, \bx, \cR) ) 
	\!\geq\! (\cW(\bx) - y)^2 
\]
and the dropout penalty is again non-negative.
Since the same calculations go through when averaging over multiple examples, 
we see that the dropout penalty is always non-negative for networks of linear nodes.
When all the weights, biases and inputs in a network of rectified linear units are positive, then the rectified linear units behave as linear units, so the dropout penalty will again be non-negative.
\qed

%% file: depends.tex
\subsection{Multi-layer dropout penalty does depend on labels}
\label{s:depends}

In contrast with its behavior on a variety of linear models including logistic
regression \cite{WWL13}, the dropout penalty can depend on the value of the response variable
in deep networks with ReLUs and the quadratic loss.
%
Thus in a fundamental and important respect, dropout differs from
traditional regularizers like weight-decay or an $L_1$ penalty.

\begin{theorem}
\label{t:depends}
There are joint distributions $P$ and $Q$, and weights
$\cW$ such that, for all dropout probabilities $q \in (0,1)$,
(a) the marginals of $P$ and $Q$ on the input variables
are equal, but (b) the dropout penalties of $\cW$ with respect to
$P$ and $Q$ are different.
\end{theorem}

We will prove Theorem~\ref{t:depends} by describing a general,
somewhat technical, condition that implies that $P$ and $Q$
are witnesses to Theorem~\ref{t:depends}.  

\newcommand{\cH}{{\cal H}}

For each input $\bx$ and dropout pattern $\cR$, let
$\underline{\cH(\cW, \bx, \cR)}$ be the values presented to the
output node with dropout.
As before, let $\underline{\bw } \in \R^n$ be those
weights of $\cW$ on connections directly into the output node
and let $b$ be the bias at the output node.
Let
$\underline{\br} \in \{ 0, 1 \}^n$ be the indicator variables for whether the
various nodes connecting to the output node are kept.  


{\bf Proof} (of Theorem~\ref{t:depends}):  Suppose that 
$P$ is concentrated on a
single $(\bx,y)$ pair.  We will then get $Q$ by modifying $y$.

Let $\bh$ be the values coming into the output node in the non-dropped out network.
Therefore the output of the non-dropout network is $\bw \cdot \bh + b$ while the output of the network with dropout is $\bw \cdot \cH(\cW, \bx, \cR)+b$.
We now examine the dropout penalty, which is the expected dropout loss minus the non-dropout loss.
We will use $\delta$ as a shorthand for $\bw \cdot (\cH(\cW, \bx, \cR) - \bh)$.
\begin{align*}
\text{dropout penalty} 
	&= \E \left( (\bw \cdot \cH(\cW, \bx, \cR) + b -y    )^2 \right) - (\bw \cdot \bh + b - y)^2 \\
 	&= \E \left( (\bw \cdot \cH(\cW, \bx, \cR) + b - \bw\cdot \bh + \bw \cdot \bh  -y )^2 \right) - (\bw \cdot \bh +b - y)^2 \\
	&= \E \left( \delta^2 \right) + 2 (\bw \cdot \bh + b -  y) \E (\delta ) 
\end{align*}
which depends on the label $y$ unless $ \E (\delta ) = 0$.


Typically $\E(\delta) \neq 0$.  
%
To prove the theorem, consider 
the case where $\bx = (1, -2)$ and there are two hidden
nodes with weights $1$ on their connections to the output node.  
The value at the hidden nodes without dropout is $0$, but with dropout
the hidden nodes never negative and computes positive values when only the negative input is dropped,
so the expectation of $\delta$ is positive. \qed





%% file: experiments.tex
\section{Experiments}
\label{s:experiments}

We did simulation experiments using Torch \cite{Tor16}.  The code is accessible at
\begin{verbatim}
https://www.dropbox.com/sh/6s2lcfrq17zshmp/AAAQ06uDa4gOAuAnw2MAghEMa?dl=0
\end{verbatim}

\subsection{Negative Weights}

We trained networks with the standard architecture with $K=5$ inputs,
depth $3$, and width $50$ on the training data studied in
Section~\ref{s:negative.weights}: one input with $\bx = (0,0,...,0)$
and $y = 0$, and one input with $\bx = (1,1,...,1)$ and $y = 1$.  We
used the \verb+nn.StochasticGradient+ function from Torch with a maximum of
$10000$ iterations, and a learning rate of
$\frac{0.1}{1 + 0.1 \times t}$ on iteration $t$.

With the above parameters, we repeated the following experiment
$1000$ times.
\begin{itemize}
\item Initialize the parameters of a network ${\cal W}_D$.
\item Clone this network, to produce a copy ${\cal W}$ with the same
initialization.
\item Train ${\cal W}_D$ with dropout probability $1/2$ at all nodes,
and train ${\cal W}$ without dropout (without using weight decay for either one).
\item Compare the number of negative weights $\mathrm{neg}({\cal W}_D)$ in
$\cW_D$ with $\mathrm{neg}(\cW)$.
\end{itemize}
Counts of the outcomes are shown in the following table.
\begin{center}
\begin{tabular}{|c|c|c|} \hline
$ \mathrm{neg}({\cal W}_D) > \mathrm{neg}(\cW) $ & $ \mathrm{neg}({\cal W}_D) < \mathrm{neg}(\cW) $ & $ \mathrm{neg}({\cal W}_D) = \mathrm{neg}(\cW) $ \\ \hline
584 & 338 & 78 \\ \hline
\end{tabular}
\end{center}

Recall that
Theorem~\ref{t:negative_weights} shows that a global optimizer of
the dropout criterion definitely uses negative weights when trained on data sources
similar to $P_{({\bf 1}, 1)}$, despite the fact that that they are consistent with
a monotone function.  
This experiment shows that even when optimization
is done in a standard way, using dropout 
tends to create models with more negative weights for $P_{({\bf 1}, 1)}$,
and we attribute this to the variance reduction effects associated with Theorem~\ref{t:negative_weights}.

The experiment also indicates that training with dropout sometimes produces fewer negative weights.
Due to the random initialization, a hidden ReLU unit
with negative weights can evaluate to $0$ on both inputs, so
its weights will never be updated without dropout.
On the other hand, the extra variance from dropout could cause updates to these negative weights.
Another way the non-dropout training could produce more negative weights is if a hidden node whose output is too large has many small weights
that turn negative with a standard gradient descent step.  However,  with dropout only about half the small weights will be updated.
Theorem~\ref{t:negative_weights} focusses on the effect of dropout at the
global minimum, and abstracts away these kinds of initialization and optimization effects.

\subsection{Scale (in)sensitivity}
\label{s:exp.scale}

Our experiments regarding the sensitivity with respect to the scale of
the input used the standard architecture with $K=5$, $d=2$, $n=5$.
We used stochastic gradient using the \verb+optim+ package for
Torch, with learning rate $\frac{0.01}{1 + 0.00001 t}$ and momentum
of 0.5, and a maximum of 100000 iterations.

We performed $10$ sets of training runs.  In each run:
\begin{itemize}
\item Ten training examples were generated uniformly at random from $[-1,1]^K$.
\item Target outputs were assigned using $y = \prod_i \sign(x_i)$.
\item Five training sets $S_1, ..., S_5$ with ten examples each were obtained by rescaling the inputs by
$\{ 0.5, 0.75, 1, 1.25, 1.5 \}$ and leaving the outputs unchanged.
\item The weights of a network $\cW_{\mathrm{init}}$ were initialized using the default initialization from Torch.
\item For each $S_i$:
  \begin{itemize}
  \item $\cW_{\mathrm{init}}$ was cloned three times to produce
    $\cW_D$, $\cW_2$ and $\cW_{\mathrm{none}}$ with identical starting parameters.
  \item $\cW_D$ was trained with dropout probability $1/2$ and no weight decay.
  \item $\cW_2$ was trained with
    weight decay with $\lambda = 1/2$ and no dropout.
    \item $\cW_{\mathrm{none}}$ was trained without any regularization.
  \end{itemize}
\end{itemize}

\begin{figure}
\begin{center}
\includegraphics[width=3in]{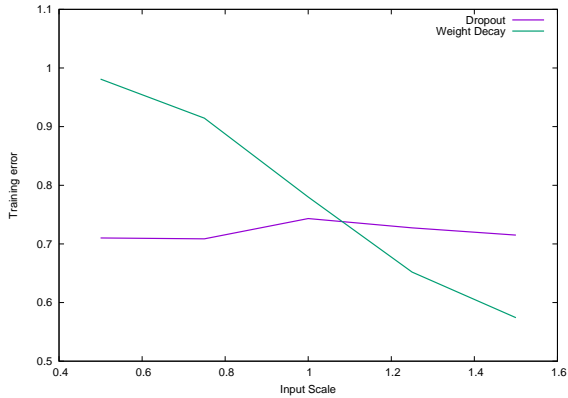} \\
\end{center}
\caption{Training error as a function of the scale of the inputs
for Dropout and Weight Decay in the experiment of Section~\ref{s:exp.scale}.}
\label{f:scale}
\end{figure}

The average training losses of $\cW_D$ and $\cW_2$, over the 10
runs, are shown in Figure~\ref{f:scale}.  (The average training loss
of $\cW_{\mathrm{none}}$ was less than $0.05$ at all scales.)

The theoretical insensitivity of dropout to the scale of the inputs
described in Theorem~\ref{t:scale-free} is also seen here, along with
the contrast with weight decay analyzed in
Theorem~\ref{t:not-scale-free}.

The scale of the inputs also affects the dynamics of stochastic
gradient descent.  With very small inputs, convergence is very slow,
and with very large inputs, SGD is unstable.  The effects of
the scale of the inputs on inductive bias analyzed in this paper are
visible at the scales where optimization can be done effectively.

%% file: conclusions.tex
\section{Conclusions} 
\label{s:conclusions}

The reasons behind dropout's surprisingly good performance in training deep networks across a variety of applications 
are somewhat mysterious
and there is relatively little existing 
formal analysis.  
A variety of explanations have been offered (e.g. \cite{BAP14,BS14,BS13,Gal2015DropoutB,WFWL14}),  including 
the possibility that dropout reduces the amount of coadaptation in a network's weights \cite{HSKSS12}.  

The dropout criterion is an expected loss over dropout patterns, 
and the variance in the output values over dropout patterns contributes to this expected loss.
Therefore dropout may co-adapt weights in order to reduce this (artificial) variance.
We prove that this happens even in very simple situations where 
nothing in the training data justifies negative weights
(Theorem~\ref{t:negative_weights}).
This indicates that the relationship between dropout and co-adaption is not a simple one.

The effects of dropout in deep neural networks are rather complicated, and approximations can be misleading
since the dropout penalty is very non-convex even in 1-layer networks~\cite{HL15}.
In Section~\ref{s:scale} we show that dropout does enjoy several scale-invariance properties that 
are not shared by weight-decay.  
A perhaps surprising consequence of these invariances is that there are never isolated local minima when learning a deep network with dropout.
Further exploration of these scale invariance properties is warranted to see if they are a contributor to dropout's empirical success
or can be exploited to facilitate training.
While contrasting dropout to weight-decay in simple situations, we found that 
a degenerate all-zero network results (Theorem~\ref{t:not-scale-free})
when the $L_2$ regularization parameter is above a threshold.
This is in dramatic contrast to our previous intuition from the 1-layer case.

In \cite{WWL13}, dropout was viewed as a regularization method, adding a data dependent 
penalty to the empirical loss of (presumably) undesirable solutions.
Section~\ref{s:dropout.penalty} shows that, unlike the generalized linear models case analyzed there,
the dropout penalty in deeper networks can be negative and depends on the labels in the training data,
and thus behaves 
unlike most regularizers.
On the other hand, the dropout penalty can grow exponentially in the depth of the network, 
and thus may better reflect the complexity of the underlying model space than $L_2$ regularization.

This paper uncovers a number of dropout's interesting fundamental properties using formal analysis of simple cases.
However, the effects of using dropout training in deep networks are subtle and complex, and we hope that this paper
lays a foundation to promote further formal analysis of dropout's properties and behavior.

%% file: notation_table.tex
\section{Table of Notation}
\label{a:notation}

\begin{tabular}{| c | l |}
\hline
Notation & Meaning \\ \hline
$\ind_\text{set}$		& indicator function for ``set'' \\
$(\bx, y)$				& an example with feature vector $\bx$ and label $y$ \\
$\sigma(\cdot)$			& the rectified linear unit computing  $\max(0, \cdot)$ \\
$\cW$				& an arbitrary weight setting for the network \\
$w, v$				& specific weights, often subscripted  \\
$\cW(\bx)$			& the output value produced by weight setting $\cW$ on input $\bx$ \\  \hline
$P$					& an arrelbitrary source distribution over $(\bx, y)$ pairs \\
$P_{\bx, y}$			& the source distribution concentrated on the single example $(\bx, y)$ \\
$R_P(\cW)$ 			& the risk (expected square loss) of $\cW$ under source $P$  \\ \hline
$q, p$				& probabilities that a node is dropped out ($q$) or kept ($p$) by the dropout process \\
$\cR$  				& a dropout pattern,  indicates the kept nodes \\
$\br, \bs$				& dropout patterns on subsets of the nodes \\
$\cD(\cW, \bx, \cR)$  	& Output of dropout with network weights $\cW$, input $\bx$, and dropout pattern $\cR$ \\ \hline
$J_D (\cW)$			& the dropout criterion  \\
$J_2 (\cW)$			& the $L_2$ criterion					\\
$\lambda$ 			& the $L_2$ regularization strength parameter  \\
$\cW_D$				& an optimizer of the dropout criterion \\
$\cW_{L_2}$			& an optimizer of the $L_2$ criterion \\ \hline
$n, d$				& the network width and depth	 			\\
$K$					& the number of input nodes  \\ \hline
\end{tabular}

%% file: elltwo.tex
\section{Proof of Theorem~\protect\ref{t:not-scale-free}}
\label{s:not-scale-free}

Here we prove Theorem~\ref{t:not-scale-free}, showing that the
weight-decay aversion depends on the values of the inputs and the
number of input nodes $K$.
%
Furthermore, unlike the single-layer
case, the $L_2$ regularization strength has a threshold where the
minimizer of the $L_2$ criterion degenerates to the all-zero network.

We will focus on the standard architecture with depth $d=2$.  
Recall that we are analyzing the distribution
$P_{(\bx,1)}$ that assigns probability $1/2$ to $(\bx,1)$
and probability $1/2$ to $({\bf 0}, 0)$.

Also, recall that, for $P_{(\bx,1)}$,
the weight-decay aversion is the maximum risk
incurred by a minimizer of $J_2$.  This motivates the following
definition.

\begin{definition}
An \underline{aversion witness} is a minimizer of 
$J_2$ that is also a maximizer of the risk $\cR$, from
among minimizers of $J_2$.
\end{definition}

The proof of Theorem~\ref{t:not-scale-free} involves a series of lemmas.

We first 
show that there is a aversion witness $\cW_{L_2}$ with a special form, and then
relate any hidden node's  effect on the output to the regularization penalty
on the weights in and out of that node. 
This will allow us to treat optimizing the $L_2$ criterion as a one-dimensional problem, whose solution yields the theorem.

For some minimizer $\cW_{L_2}$
of $J_2$, let \underline{$\bv^*_{j}$}
denote the vector of weights into hidden node $j$ and
\underline{$w^*_j$} denote the weight from $j$ to the output node.
Let $\underline{a^*_j}$ be the bias for hidden node $j$ and let
$\underline{b^*}$ be the bias for the output node.  Let
$\underline{h_j}$ be the function computed by hidden node $j$.

Note that, if there is an aversion witness with a certain property,
then we may assume without loss of generality that $\cW_{L_2}$ has
that property.


\begin{lemma}
\label{l:one.zero}
We may assume without loss of generality that for each
hidden node $j$, there is an input $\tilde \bx  \in \{ {\bf 0}, \bx \}$
such that $h_j(\tilde \bx ) = 0$.
\end{lemma}
{\bf Proof:}  Suppose neither of $h_j({\bf 0})$ or 
$h_j(\bx)$ was $0$.  If $w_j^* = 0$, then replacing both with $0$
does not affect the output of $\cW_{L_2}$, and does not increase the
penalty.  If $w_j^* \neq 0$, then subtracting $\min \{ h_j({\bf 0}), h_j(\bx) \}$
from $a_j^*$ and adding $\frac{\min \{ h_j({\bf 0}), h_j(\bx) \}}{w_j^*}$ to
$b$ does not affect $\cW_{L_2}$ or the penalty, but, after this transformation,
$\min \{ h_j({\bf 0}), h_j(\bx) \} = 0$.   
\qed

\begin{lemma}
\label{l:exact}
We may assume without loss of generality that for each
hidden node $j$, we have $| \bv_j^* \cdot \bx | = \max \{ h_j({\bf 0}), h_j(\bx) \}$.
\end{lemma}
{\bf Proof:}  
If $h_j(\bx) \geq h_j({\bf 0}) = 0$, then $a_j^* = 0$, and 
$h_j(\bx) = \bv_j^* \cdot \bx$. 

If $h_j({\bf 0}) > h_j(\bx) = 0$.
Then, $a_j^* > 0$, and $\bv_j^* \cdot \bx \leq - a_j^*$.
If needed, the magnitude of $\bv_j^*$ can be decreased to make $\bv_j^* \cdot \bx = - a_j^*$.
This decrease does not affect $\cW_{L_2}(\bx)$ or $\cW_{L_2} ( {\bf 0} )$, and can only reduce the $L_2$ penalty.
\qed

\begin{lemma}
\label{l:hidden.monotone}
We may assume without loss of generality that for each
hidden node $j$, we have $h_j({\bf 0}) = 0$.
\end{lemma}
{\bf Proof:} Suppose $h_j({\bf 0}) > 0$.  Let $z$ be this old value of
$h_j({\bf 0})$.  Then $h_j(\bx) = 0$ 
and $z = h_j({\bf 0}) = - \bv_j^* \cdot \bx$.  
If we negate $\bv^*$ and set $a_j = 0$, then
Lemma~\ref{l:exact} implies that we swap the values of $h_j(\bx)$
and $h_j({\bf 0})$.

Then, by adding $z w_j^*$ to $b^*$ and negating $w_j^*$, we correct for this swap at the output node
and do not affect
the function computed by $\cW_{L_2}$ or the penalty. \qed

Note that Lemma~\ref{l:hidden.monotone} implies that
$a_1^* = ... = a_K^* = 0$.

\begin{lemma}
\label{l:linear}
For all $j$, $\bv_j^* \cdot \bx \geq 0$.
\end{lemma}
{\bf Proof:} Since $a_j^* = 0$, if $\bv_j^* \cdot \bx < 0$, we could make
$\cW_{L_2}$ compute the same function with a smaller penalty by replacing
$\bv_j^*$ with ${\bf 0}$.
\qed

Lemma~\ref{l:linear} implies that the optimal $\cW_{L_2}$ computes the
linear function,
\begin{equation}
\cW_{L_2}(\tilde \bx) =   (\bw^*)^T V^*  \tilde \bx + b^*.
\end{equation}
Later we will call $(\bw^*)^T V^*  \tilde \bx$ the \underline{activation} at the output node.

\begin{lemma}
\label{l:b}
$b^* = \frac{ 1 - (\bw^*)^T V^* \bx }{2}$.
\end{lemma}
{\bf Proof:}  Minimize $J_2$ (wrt distribution $P_{(\bx,1)}$) as a function of $b$ using Calculus. \qed

Now, we have
\begin{equation}
\label{e:b.gone}
\cW_{L_2}(\tilde \bx ) =   \frac{1}{2} + (\bw^*)^T V^* (\tilde \bx  - \bx/2)
\end{equation}
which immediately implies
\begin{equation}
\label{e:complementary}
\cW_{L_2}({\bf 0}) =   1 - \cW_{L_2}(\bx).
\end{equation}

\begin{lemma}
\label{l:v.constant}
Each $\bv_j^*$ 
is 
a rescaling of $\bx$.
\end{lemma}
{\bf Proof:}  Projecting $\bv_j^*$ onto the span of 
$\bx$ does not affect $h_j$, and cannot increase the penalty. \qed

\begin{lemma}
\label{l:underestimate}
$\cW_{L_2}(\bx) \leq 1$.
\end{lemma}
{\bf Proof:}   By (\ref{e:complementary}),
if $\cW_{L_2}(\bx) > 1$ then 
$\cW_{L_2}({\bf 0}) < 0$ and the loss and the penalty would both
be reduced by scaling down $\bw^*$. \qed


\begin{lemma}
\label{l:max}
$\cW_{L_2}$ maximizes $\cW( \bx )$ 
over those weight vectors $\cW$ that 
have the same penalty as $\cW_{L_2}$
and compute a function of the form $\cW(\tilde x) =  \frac{1}{2} + (\bw)^T V (\tilde \bx  - \bx/2)$
(i.e.~Equation~\ref{e:b.gone}).
\end{lemma}
{\bf Proof:} 
Let $\cW$ maximize $\cW(\bx)$ over the networks considered.
If $\cW_{L_2}(  \bx  ) < \cW(  \bx  ) \leq 1$,
then $\cW$ would have the same penalty as $\cW_{L_2}$ but 
smaller error, contradicting the optimality of $\cW_{L_2}$.

If  $\cW(\bx) >  1$, then the network 
$\widetilde{\cW}$ obtained by 
scaling down the weights in the output layer 
so that  $\widetilde{\cW}( \bx ) = 1$
has a smaller penalty than  $\cW_{L_2}$ and
smaller error, again contradicting $\cW_{L_2}$'s optimality.
\qed

Informally, Lemmas~\ref{l:underestimate} and \ref{l:max} engender a
view of the learner straining against the yolk of the $L_2$ penalty to
produce a large enough output on $ \bx $.  This motivates us to ask
how large $\cW( \bx )$ can be, for a given value of $|| \cW ||_2^2$
(recall that Lemma~\ref{l:linear} allows us to assume that the biases at the hidden nodes are all 0).

\begin{definition}
For each hidden node $j$, let \underline{$\alpha_j$} be the constant such $\bv_j^* = \alpha_j \bx$, 
so that $h_j (\bx) = \alpha_j \bx \cdot \bx$.
\end{definition}

Recall that the \emph{activation} at the output node on input $\bx$ is the weighted sum of the hidden-node outputs, 
$(\bw^*)^T V^* \bx$.

\begin{definition}
The \underline{contribution to the activation} at the output due to hidden node $j$ is 
  $w_j^* h_j(\bx) = w_j^* \alpha_j \bx \cdot \bx$
and the \underline{contribution to the $L_2$ penalty} from these weights is $\frac{\lambda}{2} \left( (w_{j}^*)^2  + \alpha_j^2 \bx \cdot \bx \right)$. 
\end{definition}

We now bound the contribution to the activation in terms of the
contribution to the $L_2$ penalty.  Note that as the $L_2$ ``budget''
increases, so does the the maximum possible contribution to the output
node's activation.

\begin{lemma}
\label{l:activation.by.weight}
If $B$ is hidden node $j$'s weight-decay contribution, $(w_j^*)^2 + \alpha_j^2 \bx \cdot \bx$,
then hidden node $k$'s contribution to the output node's activation is maximized 
when
$w_{j}^* = \sqrt{\frac{B}{2}}$ and $\alpha_j  =  \sqrt{\frac{B}{2 \bx \cdot \bx} } $,  where it achieves the value
$
 B \sqrt{ \bx \cdot \bx}  / 2
$
\end{lemma}

{\bf Proof:}
Since $\alpha_j^2 \bx \cdot \bx   + (w_j^*)^2 = B$, we have $w_j^* = \sqrt{B - \alpha_j^2 \bx \cdot \bx}$,
so the contribution to the activation can be re-written as
$
\alpha_j \bx \cdot \bx \sqrt{B - \alpha_j^2 \bx \cdot \bx}.
$
Taking the derivative with respect to $\alpha_j$, and solving, we get
$\alpha_j = \pm \sqrt{\frac{B}{2 \bx \cdot \bx}}$
and we want the positive solution (otherwise the node outputs 0).
When  $\alpha_j =  \sqrt{\frac{B}{2 \bx \cdot \bx}}$ 
we have $w_j^* = \sqrt{ \frac{B}{2} }$ and
thus the node's maximum contribution to the activation is
\[
 \sqrt{\frac{ B }{ 2}}   \sqrt{\frac{B}{2 \bx \cdot \bx}} \: ( \bx \cdot \bx ) = \frac{ B \sqrt{ \bx \cdot \bx} }{ 2 }.
\]
\qed


\begin{lemma}
\label{l:how.big}
The minimum sum-squared weights for a network ${\cal W}$ (without biases at the hidden nodes) that 
has an activation $A$ at the output node on input $\bx$ is $\frac{2 A}{ \sqrt{\bx \cdot \bx} }$.
\end{lemma}
\textbf{  Proof:} When maximized, the contribution of each hidden node to
the activation at the output is $ \sqrt{ \bx \cdot \bx } /  2$ times
the hidden node's contribution to the sum of squared-weights.
Since each weight in $\cal W$ is used in exactly one hidden node's
contribution to the output node's activation, this completes the proof. \qed

Note that this bound is independent of $n$, the number of hidden units, but does depend on the input $\bx$.


\medskip

\noindent 
\textbf{  Proof (of Theorem~\ref{t:not-scale-free}):}
Let $A \geq 0$ be the activation at the output node for $\cW_{L_2}$ on input $\bx$.
From Lemma~\ref{l:b} we get that $b^* = \frac{1-A}{2}$.
Combining Lemmas~\ref{l:underestimate} , \ref{l:max}
and \ref{l:how.big}, we can re-write the $J_2$ criterion for $\cW_{L_2}$ 
and distribution $P_{ (\bx, 1) }$ in terms of $A$ as follows.  
\begin{align}
J_2 (\cW_{L_2} ) &= 
	\frac{1}{2} \left( b^* - 0 \right)^2 + \frac{1}{2} \left( \cW_{L_2}(\bx) - 1 \right)^2 + \frac{\lambda}{2} \Wsquare  \nonumber \\
	&= \frac{1}{2} \left(  
			\left( \frac{1 - A}{2} \right)^2 
			+ \left( \frac{1+A}{2} - 1 \right)^2
			+ \lambda \frac{ 2 A }{ \sqrt{\bx \cdot \bx} }  \right). \label{e:J2.crit}
\end{align}

Differentiating with respect to $A$, we see that the criterion is minimized when
\begin{align*}
A &= 1 - \frac{ 2 \lambda }{ \sqrt{\bx \cdot \bx} }  	&\text{when }  \frac{ 2 \lambda }{ \sqrt{\bx \cdot \bx} }  \leq 1 \\
A &= 0  							&\text{when }  \frac{ 2 \lambda }{ \sqrt{\bx \cdot \bx} }  > 1 
\end{align*}
since we assumed $A \geq 0$; 
when $A=0$ then $\cW_{L_2}$ has all zero weights with a bias of 1/2 at the output.

The risk part of~(\ref{e:J2.crit}) simplifies to 
\[
\frac{ (1-A)^2 }{4}  = \frac{ \lambda^2 }{ \bx \cdot \bx } ,
\]
so the overall the risk of an aversion witness 
is $\displaystyle \min \left\{ \frac{1}{4},  \frac{\lambda^2}{ \bx \cdot \bx }  \right\}$.
\qed